# Hadamard Response: Estimating Distributions Privately, Efficiently, and with Little Communication

Jayadev Acharya[*]
Cornell University
acharya@cornell.edu

Ziteng Sun[*]
Cornell University
zs335@cornell.edu

Huanyu Zhang[*]
Cornell University
hz388@cornell.edu

June 27, 2018

## Abstract

We study the problem of estimating $k$-ary distributions under $\varepsilon$-local differential privacy. $n$ samples are distributed across users who send privatized versions of their sample to a central server. All previously known sample optimal algorithms require linear (in $k$) communication from each user in the high privacy regime ($\varepsilon = O(1)$), and run in time that grows as $n \cdot k$, which can be prohibitive for large domain size $k$.

We propose *Hadamard Response (HR)*, a local privatization scheme that requires no shared randomness and is symmetric with respect to the users. Our scheme has order optimal sample complexity for all $\varepsilon$, a communication of at most $\log k + 2$ bits per user, and nearly linear running time of $\tilde{O}(n+k)$.

Our encoding and decoding are based on Hadamard matrices, and are simple to implement. The statistical performance relies on the coding theoretic aspects of Hadamard matrices, ie, the large Hamming distance between the rows. An efficient implementation of the algorithm using the Fast Walsh-Hadamard transform gives the computational gains.

We compare our approach with Randomized Response (RR), RAPPOR, and subset-selection mechanisms (SS), both theoretically, and experimentally. For $k = 10000$, our algorithm runs about 100x faster than SS, and RAPPOR.

## 1 Introduction

Estimating the underlying probability distribution from data samples is a quintessential statistical problem. Given samples from an unknown distribution $p$, the goal is to obtain an estimate $\hat{p}$ of $p$. The problem has a rich, and vast literature (see e.g. [6, 39, 17, 18], and many others), with the primary goal of statistical efficiency, namely minimizing the sample complexity for estimation, which is the first resource we consider.

**1. Utility.** What is the **sample complexity** of estimation?

In many applications, data contains sensitive information, and preserving the privacy of individuals is paramount. Without proper precautions, sensitive information can be inferred as evidenced by well publicized data leaks over the past decade, including de-anonymization of public health records in Massachusetts [41], de-anonymization of Netflix users [37] and de-anonymization of individuals participating in the genome wide association study [29]. On the policy side, the EU General Data Protection Regulation are now in effect, putting strict regulations on the data collection methods across the EU (visit http://www.eugdpr.org).

---

[*]This research is supported by NSF-CCF-CRII 1657471 , and a grant from Cornell University.

Private data release and computation on data has been studied in several fields, including statistics, machine learning, database theory, algorithm design, and cryptography (See e.g., [45, 14, 22, 46, 23, 42, 13]). *Differential Privacy (DP)* [24] has emerged as one of the most popular notions of privacy (see [24, 46, 26, 9, 36, 32], references therein, and the recent book [25]). DP has been adopted by several companies including Google, and Apple [21, 27].

A particularly popular privacy setting is *local differential privacy (LDP)* [45, 23], where users do not trust the data collector, and privatize their data before releasing. We study distribution estimation under LDP. Distribution estimation with privacy is an important problem. For example, understanding the drug usage habits of the entire population (the distribution) is crucial for policy design. Understanding the internet traffic distribution is important for ad-placement. In both these applications, preserving individual privacy is essential.

**2. Privacy.** How much information about a user is leaked by the scheme?

There are inherent trade-offs between utility and privacy. Sample privacy trade-offs have been recently studied for various problems, including distribution estimation [23, 31, 47, 43, 20, 35].

However, two crucial resources have not been considered in private distribution estimation, computation, and communication. In applications where the underlying dimensionality is high, or the number of samples is large, it is imperative to have computationally efficient algorithms. Internet companies collect information about user's browsing history over a large number of users and websites, and large departmental stores collect purchase statistics over a large number of users and products. In these problems, algorithms with high computational overhead are prohibitive, even if they have optimal sample complexity. There has been recent interest in computationally efficient distribution estimation in the non-private setting (see e.g., [15, 1, 33, 12, 16, 40, 3]).

**3. Computational Complexity.** What is the running time of the algorithm?

In distributed applications, communication (both with and without privacy) is critical. For example, a large fraction of internet traffic is on hand-held devices with limited uplink capacity due to limited battery power, limited uplink bandwidth, or expensive data rates. Similarly, in large scale distributed machine learning problems, communication from processors to the server is the bottleneck since local computations are fast. Communication limited distributed distribution estimation has been studied in the non-private setting(e.g., [48, 4, 19, 2, 28]).

In the context of private estimation tasks, the problem of finding the heavy hitters, and learning properties under local differential privacy under the assumption of public randomness, where the server can send communication to the clients to reduce communication from user end has received much attention recently [8, 7, 5, 30, 44, 11]. However, these algorithms require shared randomness, as well as asymmetric schemes, where each user can use a different privatization mechanism. [7] uses a Hadamard transform, but they use it to form orthogonal basis and reduce storage, which is different from us.

**4. Communication Complexity.** How many bits are communicated?

In this work, we consider discrete distribution estimation under the aforementioned four resources. We provide the first algorithm that is simultaneously sample order optimal for any privacy value, has logarithmic communication per symbol, and runs in linear time in the input and output size.

## 1.1 Organization.

In Section 2 we describe the problem set-up, in Section 2.1 and 2.2 we describe prior privatization schemes, and our results. In Section 3, we provide a family of $\varepsilon$-LDP privatization schemes. In Section 4, we specialize and design schemes that are optimal in the most interesting regime of high privacy. Finally in Section 5 we will describe how to extend these schemes to general $\varepsilon$.

## 2 Preliminaries

**Local Differential Privacy (LDP).** Suppose $x$ is a private information that takes values in a set $\mathcal{X}$ with $k$ elements (wlog let $\mathcal{X} = [k]:=\{0, 1, \ldots, k-1\}$). A privatization mechanism is a randomized mapping $Q$ from $[k]$ to an output set $\mathcal{Z}$ (which can be arbitrary), that maps $x \in \mathcal{X}$ to $z \in \mathcal{Z}$ with probability $Q(z|x)$. The output $z$ of this mapping, called the privatized sample, is then released. $Q$ is $\varepsilon$-locally differentially private ($\varepsilon$-LDP) [23] if for all $x, x' \in \mathcal{X}$,

$$\sup_{z \in \mathcal{Z}} \frac{Q(z|x)}{Q(z|x')} \leq e^{\varepsilon}. \tag{1}$$

Small values of $\varepsilon$ are more stringent and is the high privacy regime, and large values of $\varepsilon$ is the low privacy regime. When $\mathcal{X}$ and $\mathcal{Z}$ are both discrete, the mechanism $Q$ is described by a stochastic matrix of size $|\mathcal{X}| \times |\mathcal{Z}|$ whose $(x, z)$th entry is $Q(z|x)$. $Q$ is $\varepsilon$-LDP if the ratio of *any two entries* in a column of this matrix is at most $e^{\varepsilon}$.

**Randomness, and Symmetry.** A scheme that requires shared/public randomness requires the generation of shared randomness at the server, which needs to be communicated to the users. Symmetric schemes are those where each user uses the same privatization scheme [38]. In this paper, we consider schemes that are symmetric and require no shared randomness. Other such schemes include RAPPOR, Randomized Response, and subset selection methods, described later. We note that the literature on heavy hitter estimation has mostly considered schemes with shared randomness [8, 7, 11], and it will be interesting to see if our methods can provide improved algorithms for the heavy hitter problem.

**LDP distribution estimation.** Let $\Delta_k = \left\{ p(0), \ldots, p(k-1) : p(x) \geq 0, \sum_{x=0}^{k-1} p(x) = 1 \right\}$ be the set of all distributions over $[k]$. Let $X_1, \ldots, X_n$ be independent samples drawn from an *unknown* $p \in \Delta_k$, where $X_i$ is the private (sensitive) data with the $i$th user. Each user maps $X_i$ through an $\varepsilon$-LDP $Q$, to obtain $Z_i$. The task at the server, upon observing the privatized samples $Z_1, \ldots, Z_n$, is to output $\hat{p} : \mathcal{Z}^n \to \Delta_k$, an estimate of $p$. Let $d : \Delta_k \times \Delta_k \to \mathbb{R}_+$ be a distance measure between distributions in $\Delta_k$. Private distribution estimation task is the following:

> Given $\alpha > 0, \varepsilon > 0, d : \Delta_k \times \Delta_k \to \mathbb{R}$, design an $\varepsilon$-LDP $Q$, and a corresponding estimation $\hat{p}$, such that $\forall p \in \Delta_k$, with probability at least 0.9, $d(\hat{p}, p) < \alpha$.

The *sample complexity* is the least $n$ for which such an $\varepsilon$-LDP scheme $Q$, and a corresponding $\hat{p}$ exists. The *communication complexity* is the number of bits to send $Z_i$ to the server. The *computational complexity* is the total time to estimate $\hat{p}$ from $Z_1, \ldots, Z_n$ at the server and to privatize $X_i$ using $Q$ at the users.

We will use $\ell_1$, and $\ell_2$ distance in this paper. For $r \geq 0$, the $\ell_r$ distance between $p, q \in \Delta_k$ is $\ell_r(p, q) := (\sum_x |p(x) - q(x)|^r)^{1/r}$. In non-private setting, the sample complexity of distribution estimation under these distances is known even including precise constants [10, 34].

| $\varepsilon$ | $k$-RR | RAPPOR | $k$-SS | $\varepsilon$-HR |
|---|---|---|---|---|
| $(0, 1)$ | $\frac{k^3}{\varepsilon^2 \alpha^2}$ | $\frac{k^2}{\varepsilon^2 \alpha^2}$ | $\frac{k^2}{\varepsilon^2 \alpha^2}$ | $\frac{k^2}{\varepsilon^2 \alpha^2}$ |
| $(1, \log k)$ | $\frac{k^3}{e^{2\varepsilon}\alpha^2}$ | $\frac{k^2}{e^{\varepsilon/2}\alpha^2}$ | $\frac{k^2}{e^{\varepsilon}\alpha^2}$ | $\frac{k^2}{e^{\varepsilon}\alpha^2}$ |
| $(\log k, 2\log k)$ | $\frac{k}{\alpha^2}$ | $\frac{k^2}{e^{\varepsilon/2}\alpha^2}$ | $\frac{k}{\alpha^2}$ | $\frac{k}{\alpha^2}$ |
| $(2\log k, +\infty)$ | $\frac{k}{\alpha^2}$ | $\frac{k}{\alpha^2}$ | $\frac{k}{\alpha^2}$ | $\frac{k}{\alpha^2}$ |

Table 1: Sample complexity, up to constant factors, under $\ell_1$ distance for the different methods. The sample complexity under $\ell_2$ distance is exactly a factor $k$ smaller in each cell above.

| $\varepsilon$ | $k$-RR | RAPPOR | $k$-SS | $\varepsilon$-HR |
|---|---|---|---|---|
| $(0, 1)$ | $\log k$ | $k$ | $k$ | $\log k$ |
| $(1, \log k)$ | $\log k$ | $\frac{k}{e^{\varepsilon/2}}$ | $\frac{k}{e^{\varepsilon}}$ | $\log k$ |
| $(\log k, 2\log k)$ | $\log k$ | $\frac{k}{e^{\varepsilon/2}}$ | $\log k$ | $\log k$ |
| $(2\log k, +\infty)$ | $\log k$ | $\log k$ | $\log k$ | $\log k$ |

Table 2: Communication requirements for distribution estimation techniques.

## 2.1 The privatization mechanisms

We will now briefly describe RR, RAPPOR, the most popular $\varepsilon$-LDP schemes using no interaction and public randomness. We will also mention SS, and our proposed HR. For a detailed description of RAPPOR and SS, please refer to Section C.

**$k$-Randomized Response (RR).** The $k$-RR mechanism [45, 31] is an $\varepsilon$-LDP $Q_{\text{RR}}$ with $\mathcal{Z} = \mathcal{X} = [k]$, such that

$$Q_{\text{RR}}(z|x) := \begin{cases} \frac{e^{\varepsilon}}{e^{\varepsilon}+k-1} & \text{if } z = x, \\ \frac{1}{e^{\varepsilon}+k-1} & \text{otherwise.} \end{cases} \quad (2)$$

**$k$-RAPPOR.** The randomized aggregatable privacy-preserving ordinal response (RAPPOR) is an $\varepsilon$-LDP mechanism which was proposed in [23, 27]. Its simplest implementation $k$-RAPPOR maps $\mathcal{X} = [k]$ to $\mathcal{Z} = \{0, 1\}^k$. It first does a one hot encoding to the input $x \in [k]$ to obtain $\mathbf{y} \in \{0, 1\}^k$, such that $\mathbf{y}_j = 1$ for $j = x$, and $\mathbf{y}_j = 0$ for $j \neq x$. The privatized output of $k$-RAPPOR is a $k$-bit vector obtained by independently flipping each bit of $\mathbf{y}$ with probability $\frac{1}{e^{\varepsilon/2}+1}$.

**Subset Selection techniques.** [43, 47] propose an $\varepsilon$-LDP scheme that maps $x \in [k]$ to subsets of $[k]$ of size $\lceil k/(e^{\varepsilon}+1) \rceil$. The scheme is described in detail in Section C.

**Hadamard Response.** We propose Hadamard Response (HR), an $\varepsilon$-LDP scheme with $\mathcal{Z} = [K]$, for some $k \leq K \leq 4k$. The algorithm is described in Section 4 for high privacy, and in Section 5 for general privacy.

## 2.2 Previous Results

To estimate distributions in $\Delta_k$ to $\ell_1$ distance $\alpha$ under $\varepsilon$-LDP, the sample, communication and time requirements of the various schemes are given in Table 1, 2 and 3 respectively..

| $k$-RR | $k$-RAPPOR | Subset selection | $\varepsilon$-HR |
|--------|------------|------------------|------------------|
| $n+k$  | $n+k+\frac{nk}{e^{\varepsilon/2}}$ | $n+k+\frac{nk}{e^\varepsilon}$ | $n+k$ |

Table 3: Time bounds for distribution estimation. The running times are described in Section C. These are upper bounds up to logarithmic factors.

The sample complexity is given in **Table 1**. The entries in *green* boxes are sample-order optimal, namely there is a matching lower bound [47]. Note that RR is sample-optimal in the low privacy regime (last two rows), and is *highly sub-optimal* in the high privacy regime ($\varepsilon = O(1)$). RAPPOR is optimal for high-privacy, but sub-optimal for medium privacy. SS, and our proposed HR are sample-order-optimal for all $\varepsilon$. The sample complexity arguments for RR, RAPPOR, and SS can be found in [31, 47].

**Table 2** describes the communication requirements of various schemes. However, it is not clear how to measure the communication requirements, since for a given privatization scheme, there might be communication protocols requiring fewer bits than others. For example, RAPPOR is described as giving $k$ bits as its output, but perhaps these $k$ bits can be compressed further requiring much smaller communication. We get around such concerns by observing that, once the input distribution $p$ and the privatization mechanism $Q$ is fixed, the output distribution of the privatized sample $Z$ is fixed. By Shannon's source coding theorem, to *faithfully* send $Z$ to the server requires at least $H(Z)$ bits of communication. The entries in the table are derived by considering the input distribution to be near uniform, and evaluating the entropy of the output of the mechanisms. For RR, $\log k$ bits of communication follows from $\mathcal{Z} = [k]$. Note that in this paper all logarithms are in base 2. The communication requirements for RAPPOR, and SS are derived in Section C (Theorems 9, and Theorem 10 respectively).

**Table 3** describes the total running time lower bounds for faithfully implementing the known schemes. The argument is that at the server, the computation complexity is at least the number of bits that need to be read, which is the amount of communication from the users. If there are $n$ users, then $n \cdot H(Z)$ serves as our time complexity bound, and these form the entries in the table.

### 2.3 Motivation and Our Results

Our work is motivated by the first three columns of the tables, which captures the apparent sample-communication-computation trade-offs present in the existing schemes. We elaborate this point in the most interesting regime of high privacy. For simplicity, fix $\varepsilon = 1$, and $\alpha = 0.1$ (chosen arbitrarily!), and treat them as fixed constants in this paragraph. In this setting, from Table 1, note that the optimal sample complexity is $\Theta(k^2)$, achieved by RAPPOR, and SS, while RR has a sub-optimal sample complexity of $\Theta(k^3)$. Now consider the communication requirements. $\mathcal{Z} = [k]$ for RR, requiring only $\log k$ bits. A straight-forward computation shows that any input distribution to the RAPPOR mechanism induces an output distribution over $\{0,1\}^k$ with entropy at least $\Omega(k)$, thus requiring $\Omega(k)$ bits to *faithfully* send the privatized samples to the server. SS also requires $\Omega(k)$ bits in this regime. These are formally shown in Theorem 9 and Theorem 10. As for the running time at the server end, a bound of $\Omega(k^3)$ for all these three methods follows from the total communication to the server (#samples × #bits per sample), which is a factor $k$ larger than the $\Theta(k^2)$ optimal sample complexity bound.

Our main result is the following, which is formally stated in Theorem 2, and Theorem 7.

**Theorem 1.** *We propose a simple algorithm for $\varepsilon$-LDP distribution estimation that for all param-*

eter regimes, is sample optimal, runs in near-linear time in the number of samples, and has only a logarithmic communication complexity in the domain size, for both the $\ell_1$, and $\ell_2$ distance.

Going back to the high privacy regime, considered before, this shows that our scheme has a running time of $\tilde{O}(k^2)$, which is nearly linear in the optimal sample complexity under $\ell_1$ distance.

## 3 A family of $\varepsilon$-LDP schemes

We first propose a general family of LDP schemes, and then carefully choose schemes from this family that are sample-optimal, communication and computationally efficient for distribution estimation.

The scheme involves the following steps:

1. Choose an integer $K$, and let the output alphabet be $\mathcal{Z} = [K]$.
2. Choose a positive integer $s \leq K$.
3. For each $x \in \mathcal{X} = [k]$, pick $C_x \subseteq [K]$ with $|C_x| = s$.
4. The privatization scheme from $[k]$ to $[K]$ is then given by:

$$Q(z|x) := \begin{cases} \frac{e^\varepsilon}{se^\varepsilon + K - s} & \text{if } z \in C_x, \\ \frac{1}{se^\varepsilon + K - s} & \text{if } z \in \mathcal{Z} \setminus C_x. \end{cases} \tag{3}$$

This scheme satisfies (1), and is $\varepsilon$-LDP. This privatization scheme chooses a set $C_x$ for each $x$ and assigns the elements in $C_x$ a higher probability than those not in $C_x$. We also note that RR is a special case of this construction when $K = k$, $s = 1$, and $C_x = \{x\}$. We know from the last section that RR is sub-optimal in the high privacy regime. Our general inspiration comes from coding theory, and we select $s$, and $C_x$ carefully in order to send more information across $Q$ than RR.

In Section 4 we give an optimal scheme in the high privacy regime, and extend it to the general case in Section 5

## 4 Optimal scheme for high privacy regime

**Privatization scheme.** If for two $x$, and $x'$, $C_x = C_{x'}$, then we cannot tell them apart. Therefore, the hope is that the farther apart $C_x$ and $C_{x'}$ are, the easier it is to tell them apart. With this in mind, we specify a particular choice of parameters for our scheme, which turns out to be sample-optimal in the high privacy regime. In particular, our privatization scheme will satisfy the following:

---

**An optimal privatization for high privacy**

Choose $K$, and $C_x$'s such that (We will show in Section 4.1 how to satisfy these conditions.):

**C1.** $K$ is between $k$ and $2k$, and $s = K/2$, namely for all $x \in [k]$, $|C_x| = \frac{K}{2}$.

**C2.** For any $x, x' \in [k]$, and $x \neq x'$, $|\Delta(C_x, C_{x'})| = |(C_x \setminus C_{x'}) \cup (C_{x'} \setminus C_x)| = \frac{K}{2}$.

Use (3) for privatization.

---

**Performance.** We will show that for $\varepsilon = O(1)$, this privatization is sample-order-optimal, namely there is a corresponding estimator $\hat{p} : [K]^n \to \Delta_k$ that is sample-optimal. Before describing the estimation procedure, we provide the statistical guarantees.

**Theorem 2.** *For any privatization scheme satisfying **C1, C2**, there is a corresponding estimation scheme $\hat{p} : [K]^n \to \Delta_k$, such that*

$$\mathbb{E}\left[\ell_2^2(\hat{p}, p)\right] \leq \frac{4k(e^\varepsilon + 1)^2}{n(e^\varepsilon - 1)^2}, \text{ and } \mathbb{E}\left[\ell_1(\hat{p}, p)\right] \leq \sqrt{\frac{4k^2(e^\varepsilon + 1)^2}{n(e^\varepsilon - 1)^2}}. \tag{4}$$

The sample optimality, and small communication for high privacy is an immediate corollary.

**Corollary 3.** *When $\varepsilon = O(1)$, the sample complexity of this scheme for estimation to $\ell_1$ distance $\alpha$ is $O(k^2/\varepsilon^2 \alpha^2)$ samples, and for $\ell_2^2$ distance is $O(k/\varepsilon^2 \alpha^2)$. Further, the communication from each user is at most $\log(k) + 1$ bits. This is sample-optimal for $\varepsilon = O(1)$ for both $\ell_1$ (Table 1) and $\ell_2^2$ (see [47]).*

*Proof.* Applying Markov's inequality in Theorem 2, and substituting $e^\varepsilon + 1 = \Theta(1)$, and $e^\varepsilon - 1 = \Theta(\varepsilon)$ when $\varepsilon = O(1)$ gives the sample complexity bounds. The communication bounds are from $\log K \leq \log(k) + 1$. □

**Estimation.** Suppose $Q_{K,\varepsilon}$ is an $\varepsilon$-LDP scheme satisfying **C1**, and **C2**. For an input distribution $p$ over $[k]$, let $p(C_x)$ be the probability that the privatized sample $Z \in C_x$. Using $|C_x| = K/2$, and **C2**, it follows that $|C_x \setminus C_{x'}| = K/4$, and $|C_x \cap C_{x'}| = K/4$. Therefore,

$$p(C_x) = p(x)\left(\sum_{z \in C_x} Q_{K,\varepsilon}(z|x)\right) + \sum_{x' \neq x} p(x')\left(\sum_{z \in C_x \setminus C_{x'}} Q_{K,\varepsilon}(z|x') + \sum_{z \in C_x \cap C_{x'}} Q_{K,\varepsilon}(z|x')\right)$$

$$= p(x)|C_x|\frac{e^\varepsilon}{(se^\varepsilon + K - s)} + \sum_{x' \neq x} p(x')\left(\frac{|C_x \setminus C_{x'}| \cdot 1}{se^\varepsilon + K - s} + \frac{|C_x \cap C_{x'}| \cdot e^\varepsilon}{se^\varepsilon + K - s}\right) \tag{5}$$

$$= \frac{1}{2} + \frac{e^\varepsilon - 1}{2(e^\varepsilon + 1)}p(x), \tag{6}$$

where (5) follows from (3), and (6) by plugging $s = K/2$, and from **C2**. We can rewrite this as

$$p(x) = \frac{2(e^\varepsilon + 1)}{e^\varepsilon - 1}\left(p(C_x) - \frac{1}{2}\right). \tag{7}$$

This forms the basis of our estimation. From the privatized samples, we estimate of $p(C_x)$, and from that we estimate $p$. The entire scheme is given below.

---

**An optimal distribution estimation scheme for high privacy**

**Input:** $k$, $\varepsilon$, privatized samples $Z_1, \ldots, Z_n$

1. For each $x \in [k]$, estimate $p(C_x)$ with its empirical probability:

$$\widehat{p(C_x)} := \sum_{j=1}^{n} \frac{\mathbb{I}\{Z_j \in C_x\}}{n}. \tag{8}$$

2. Estimate $\hat{p}$ as:

$$\hat{p}(x) := \frac{2(e^\varepsilon + 1)}{e^\varepsilon - 1}\left(\widehat{p(C_x)} - \frac{1}{2}\right). \tag{9}$$

---

**Proof of Theorem 2.**[1] Let $p(C), \widehat{p(C)}$, be the vector of probabilities of $p(C_x)$'s and $\widehat{p(C_x)}$'s respectively. From (7) and (9),

$$\mathbb{E}\left[\ell_2^2(\hat{p}, p)\right] = \frac{4(e^\varepsilon + 1)^2}{(e^\varepsilon - 1)^2} \mathbb{E}\left[\ell_2^2(\widehat{p(C)}, p(C))\right].$$

From (8), $\mathbb{E}\left[\widehat{p(C_x)}\right] = \mathbb{E}[\mathbb{I}\{Z_j \in C_x\}] = p(C_x)$. Therefore,

$$\mathbb{E}\left[\ell_2^2(\widehat{p(C)}, p(C))\right] = \mathbb{E}\left[\sum_{x \in [k]} (\widehat{p(C_x)} - p(C_x))^2\right] = \sum_{x \in [k]} \mathbb{E}\left[(\widehat{p(C_x)} - p(C_x))^2\right] = \sum_{x \in [k]} \operatorname{Var}(\widehat{p(C_x)}).$$

By the independence of $Z_i$'s, $\widehat{p(C_x)}$ is the average of $n$ independent Bernoulli random variables each with expectation $p(C_x)$. Hence,

$$\sum_{x \in [k]} \operatorname{Var}(\widehat{p(C_x)}) = \sum_{x \in [k]} \frac{1}{n} \cdot p(C_x)(1 - p(C_x)) \leq \frac{1}{n} \sum_{x \in [k]} p(C_x) \leq \frac{k}{n}.$$

Plugging this bound in the previous expression gives the bound on $\ell_2^2$ distance of the theorem.

$$\mathbb{E}\left[\ell_2^2(\hat{p}, p)\right] \leq \frac{4k(e^\varepsilon + 1)^2}{n(e^\varepsilon - 1)^2}. \tag{10}$$

Using $k \cdot \ell_2^2(\hat{p}, p) \geq \ell_1(\hat{p}, p)^2$ with (10) gives the desired bound on $\mathbb{E}[\ell_1(\hat{p}, p)]$. □

### 4.1 Computational complexity and Hadamard matrices.

We showed the sample, and communication complexity guarantees. However, two questions are still unanswered:
- How to choose $K$, and design $C_x$'s that satisfy **C1, C2**?
- What is the time complexity of privatization and estimation?

We now address these questions. We start with the computational requirements of the proposed scheme, assuming **C1,C2**.

**Computation at users.** Given $C_x$'s, each user needs to implement (3). This requires uniform sampling from $C_x$'s, as well as from $[K] \setminus C_x$. We will design schemes to do this in time $O(\log K)$.

**Computation at the server.** The server needs to implement (8) and (9). Note that (9) can be implemented in time $O(k)$ after implementing (8). However, a straightforward implementation of (8) requires $n \cdot k$ time, since for each $x$ we iterate over all the samples, giving running time of $O(n \cdot k)$. In particular, in the high privacy regime (say with $\varepsilon = 1$, and $\alpha = 0.1$) the sample complexity is $O(k^2)$ but the time requirement will be $O(k^3)$. We now show how to design a privatization to satisfy **C1, C2**, and for which we can implement (8) in time only $\tilde{O}(n + k)$.

**Hadamard Response (HR) for high privacy.** Suppose $K$ is a power of two, and let $H_K \in \{\pm 1\}^{K \times K}$ be the Hadamard matrix of size $K \times K$ designed by the well known Sylvester's construction as follows. Let $H_1 = [1]$, and for $m = 2^j$, for $j \geq 1$, then

$$H_m := \begin{bmatrix} H_{m/2} & H_{m/2} \\ H_{m/2} & -H_{m/2} \end{bmatrix}.$$

---
[1] A technicality here is that $\hat{p}(x)$'s can be negative, but we can project $\hat{p}$ onto the simplex with the same order performance. We therefore only analyze the performance of $\hat{p}$ described in (9).

Some standard properties of Hadamard matrices that we use are the following:

(i) The number of +1's in each row except the first is $K/2$,
(ii) Any two rows agree (and disagree) on exactly $K/2$ locations,
(iii) Vector multiplication with $H_K$ is possible in time $O(K \log K)$ with Fast Walsh Hadamard transform,
(iv) We can uniformly sample from the +1's (and the −1's) in any row in time $O(\log K)$.

We now describe the parameters for the privacy mechanism:

1. Choice of $K$: Let $K = 2^{\lceil \log_2(k+1) \rceil} \geq k + 1$, the smallest power of 2 larger than $k$. To satisfy **C1**, we will choose $s = K/2$.
2. Choice of $C_x$'s: Map the symbols $[k] = \{0, \ldots, k-1\}$ to rows of $H_K$ as follows: map 0 to the second row, 1 to the third row, and so on. In other words, $x$ is mapped to row $x + 1$. Given any $x$, we choose $C_x \subset [K]$ to be the column indices with a '+1' in the $(x+1)$th row of $H_K$.

By Property (i) and (ii) of $H_K$, both **C1**, and **C2** are satisfied. This implies a privatization scheme with optimal sample and communication complexity in the high privacy regime.

**Fast computation with HR.** By Property (iv), we can efficiently implement the privatization scheme at the users. We will now provide an efficient implementation of (8). Let $q = (q(0), \ldots, q(K-1))$ be the vector of the empirical distribution of $Z_1, \ldots, Z_n$ over $[K] = \{0, \ldots, K-1\}$, namely

$$q(z) = \sum_{i=1}^{n} \frac{\mathbb{I}\{Z_i = z\}}{n}.$$

We can compute $q$ in linear time with a single pass over $Z_1, \ldots, Z_n$. Consider the matrix vector product $\mathbf{c} = H_K \cdot q$. For $x \in [k]$, the $(x+1)$th entry of $H_K \cdot q$ is $\sum_{z=0}^{K-1} H_K(x+1, z) \cdot q(z)$. Now note that the +1's in the $(x+1)$th column correspond to $C_x$ by construction, therefore

$$\sum_{z=0}^{K-1} H_K(x+1, z) \cdot q(z) = \sum_{z \in C_x} q(z) - \sum_{z \in [K] \setminus C_x} q(z) = 2\widehat{p(C_x)} - 1 \tag{11}$$

$$= \left(\frac{e^\varepsilon - 1}{e^\varepsilon + 1}\right) \hat{p}(x), \tag{12}$$

where (11) follows from observing that $\sum_{z \in C_x} q(z) = \widehat{p(C_x)}$ from (8), and (12) follows from (9). Therefore the estimator $\hat{p}$ is simply entries of a Hadamard vector product, appropriately normalized. By property (iii), this can be done in time $O(K \log K) = O(k \log k)$. This computational advantage is captured in the following theorem:

**Theorem 4.** *HR is an $\varepsilon$-LDP mechanism satisfying Theorem 2 that has a running time $\tilde{O}(n+k)$.*

## 5 General privacy regimes

Recall that RR is optimal for the low-privacy regime, which corresponds to $s = 1$. In the high privacy regime, we used $s = O(k)$. For general $\varepsilon$'s, we propose schemes within the framework of Section 3 that interpolate between HR and RR, while achieving the optimal sample complexity for every $\varepsilon$. In general, our choice of $s$ will be close to $\max\{k/e^\varepsilon, 1\}$ as we will see below.

**Privatization scheme in general privacy regime.** We describe how we choose $K$, $s$, and $C_x$'s, in the scheme from Section 3.

We will consider block-structured matrices that interpolate between Hadamard matrices in high privacy regime and the identity matrix (corresponding to RR) in the low privacy regime.

**Definition 5.** *Let $B$ and $b$ be powers of two, and let $K = B \cdot b$. A $(B, b)$-"reduced" Hadamard matrix is a $K \times K$ matrix with entries in $\{-1, +1\}$ defined as:*

$$H_K^b := \begin{bmatrix} H_b & P_b & \cdots & P_b \\ P_b & H_b & \cdots & P_b \\ \vdots & \vdots & \ddots & \vdots \\ P_b & P_b & \cdots & H_b \end{bmatrix},$$

*where $H_b$ is the $b \times b$ Hadamard matrix, and $P_b$ is the $b \times b$ matrix with all entries '$-1$'. Note that there are $B$ occurrences of $H_b$ along the diagonal.*

1. Choice of $K$: Let $B$ be the largest power of 2 less than $\min\{e^\varepsilon, 2k\}$, and $b$ is the smallest power of 2 larger than $\frac{k}{B} + 1$, i.e.,

$$B := 2^{\lceil \log_2 \min\{e^\varepsilon, 2k\} \rceil - 1}, \quad b := 2^{\lceil \log_2(\frac{k}{B} + 1) \rceil}.$$

Let $K = B \cdot b$. A simple computation shows that $K \le 4k$, implying that the communication from the users is at most $\log k + 2$ bits.

2. We will choose $s = b/2$.

3. Choice of $C_x$: From Property (i) of Hadamard matrices in the last section, the number of $+1$'s in the rows of $H_K^b$ corresponding to the first rows of the embedded $H_b$'s is $b$, and for all other rows it is $b/2$. Similar to the high privacy regime, we map each $x$ to a distinct row $r_x$ of $H_K^b$ with $b/2$ entries as $+1$'s. A simple way to do as before would be to map $0 \in [k]$ to the second row of $H_K^b$, and $x \in [k]$ is assigned to row $r_{x-1} + 1$, if it is not the first row of an embedded $H_b$, otherwise we assign it to $r_{x-1} + 2$. As before, let $C_x$ be the column's with a $+1$ in the $r_x$th row of $H_K^b$.

4. The privatization mechanism then applies (3), which can be done in time $O(\log k)$ at each user by Property (iv) of Hadamard matrices.

**Estimation scheme in general privacy regime.** In the high privacy regime, we related $p(C_x)$ to $p(x)$ in (7). We will do the same here, however, because of the block-structure, the inputs that map to different blocks behave differently. Let $S_i \subseteq [K]$ be the columns of the $i$th embedded $H_b$ block. Similar to $p(C_x)$, let $p(S_i)$ to be the probability that the output $z \in S_i$, when the input distribution is $p$. In other words,

$$S_i := \left\{z \mid \lfloor \frac{z}{b} \rfloor = i\right\}, \quad p(S_i) := \sum_x p(x) \left(\sum_{z \in S_i} Q(z|x)\right).$$

Similar to (7), the following lemma relates $p(C_x)$, $p(S_i)$, and $p(x)$. It is proved in Section A.

**Lemma 6.** *For the input distribution $p$, and $x \in [k]$ such that $r_x$ is in the $i$th embedded $H_b$,*

$$p(C_x) - \frac{1}{2}p(S_i) = \frac{e^\varepsilon - 1}{2(2B - 1 + e^\varepsilon)} p(x). \tag{13}$$

With this lemma, our estimation algorithm is the following:

---

**Distribution estimation for general privacy**

**Input:** $k$, $\varepsilon$, privatized samples $Z_1, \ldots, Z_n$

1. For each $x \in [k]$, estimate $p(C_x)$ with its empirical probability:

$$\widehat{p(C_x)} := \sum_{j=1}^{n} \frac{\mathbb{I}\{Z_j \in C_x\}}{n}. \tag{14}$$

2. For each $i \in B$, estimate $p(S_i)$ with its empirical probability:

$$\widehat{p(S_i)} := \sum_{j=1}^{n} \frac{\mathbb{I}\{Z_j \in S_i\}}{n}. \tag{15}$$

3. The estimator $\hat{p}$ is then given by

$$\hat{p}(x) = \frac{2(2B - 1 + e^\varepsilon)}{e^\varepsilon - 1} \cdot \left( \widehat{p(C_x)} - \frac{1}{2}\widehat{p(S_i)} \right). \tag{16}$$

---

## 5.1 Performance

Our main performance bound on this scheme is given below. The analysis is similar to the high privacy regime and is given in Section B.

**Theorem 7.** *For all values of $\varepsilon$, and $k$, and the privatization scheme above, there is an estimate $\hat{p}$ such that*

$$\mathbb{E}\left[\ell_2^2(\hat{p}, p)\right] \leq \frac{36(k + (e^\varepsilon - 1)b)e^\varepsilon}{n(e^\varepsilon - 1)^2}, \quad \mathbb{E}\left[\ell_1(\hat{p}, p)\right] \leq \sqrt{36 \frac{k}{n} \frac{(k + (e^\varepsilon - 1)b)e^\varepsilon}{(e^\varepsilon - 1)^2}}.$$

*The running time is $\tilde{O}(n + k)$, and communication is at most $\log k + 2$ bits.*

**Corollary 8.** *Plugging the values of $b$ in different regimes we obtain*

$$\mathbb{E}\left[\ell_1(\hat{p}, p)\right] \leq \begin{cases} O\left(\sqrt{\frac{k^2}{n\varepsilon^2}}\right), & \text{if } \varepsilon < 1, \\ O\left(\sqrt{\frac{k^2}{ne^\varepsilon}}\right), & \text{if } 1 < \varepsilon < \log k, \\ O\left(\sqrt{\frac{k}{n}}\right), & \text{if } \varepsilon > \log k. \end{cases}$$

*Applying Markov's inequality, we obtain* all the sample complexity *bounds for HR described in the last column of Table 1.*

## 6 Experiments.

We experimentally compare our algorithm with RR, RAPPOR and SS. We set $k \in \{100, 1000, 5000, 10000\}$, $n \in \{50000, 100000, 150000, \ldots, 1000000\}$, and $\varepsilon \in \{0.1, 0.5, 1, 2, 3, 4, 5, 6, 7, 8, 9, 10\}$. We consider geometric distributions $Geo(\lambda)$, where $p(i) \propto (1-\lambda)^i \lambda$, Zipf distributions $Zipf(k,t)$ where $p(i) \propto (i+1)^{-t}$, two-step distributions, and uniform distributions. For every setting of $(k, p, n, \varepsilon)$, and for each scheme, we simulate 30 runs, and compute the averaged $\ell_1$ error, and averaged decoding time at the server. Our code and data for the experiments can be found at https://github.com/jlyx417353617/hadamard_response.

In a nutshell, we observe that in each regime, the statistical performance of HR is comparable to the best possible. Moreover, the decoding time of HR is similar to that of RR. In comparison to RAPPOR and SS, our running times can be orders of magnitude smaller, particularly for large $k$, and small $\varepsilon$. We remark that we implement RAPPOR, and SS such that their running time is almost linear in the time needed to read the already compressed communication from the users.

We describe some of our experimental results here. Figure 1 plots the $\ell_1$ error for estimating geometric distribution for $k = 1000$. Note that for $\varepsilon = 0.5$, and $\varepsilon = 7$, our performance matches with the best schemes. In all the plots SS has the best statistical performance, however that can come at the cost of higher communication, and computation. Figures 2 captures similar statistical performance results for the uniform distribution for $k = 1000$. For larger $k$ such as $k = 10000$, the performance is shown in figure 3.

The running time of our algorithm is theoretically a factor $k/\log k$ smaller than RAPPOR and subset selection. This is evident from the plots which show that for large $k$ the running times of RAPPOR and SS are orders of magnitude more than HR, and RR.

Figure 4 shows the decoding time for the algorithms when $k = 100, 1000, 5000, 10000$ and $\varepsilon = 1$. It can be seen that our algorithm is orders of magnitudes faster in comparison to $k$-RAPPOR and $k$-SS. The gap in computation gets larger when $k$ is larger, which is consistent with our theoretical analysis. For example, for $k = 10000$, our algorithm runs 100x faster than SS, and RAPPOR.

To compare the decoding time more fairly, we use a fast implementation for RAPPOR and SS in the middle and low privacy regime. We encode the $k$ bit vector into a list of locations where there is a '+1'. The decoder uses this list to compute the histogram. The time requirement is $O(\frac{nk}{1+e^\varepsilon})$ and $O(\frac{nk}{1+e^{\varepsilon/2}})$ for SS and $k$-RAPPOR respectively in expectation (a naive implementation takes $O(nk)$ time). This is the best anyone can do to faithfully implement the algorithms.

Figure 5 shows the decoding time in middle privacy regime. We use fast implementation for RAPPOR and SS here. We can see our proposed algorithm is still saving a lot of time comparing to $k$-RAPPOR and SS. For low privacy regime, essentially everything breaks down to Randomized Response, so we won't show the plots for the time comparison in this regime.

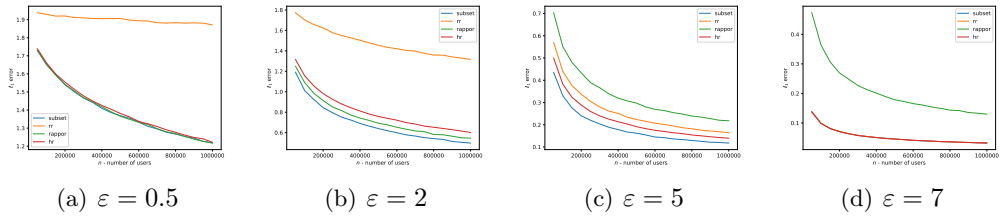

(a) $\varepsilon = 0.5$    (b) $\varepsilon = 2$    (c) $\varepsilon = 5$    (d) $\varepsilon = 7$

Figure 1: $\ell_1$-error comparison between four algorithms $k = 1000$ and $p \sim Geo(0.8)$

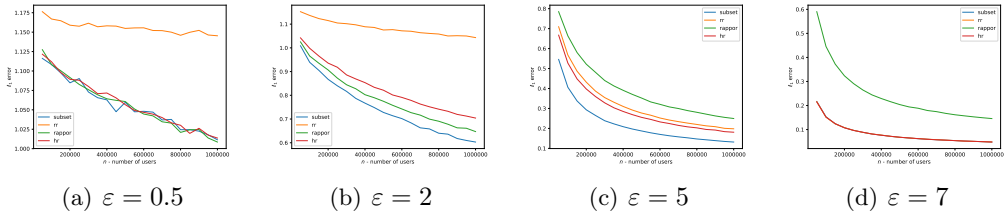

(a) $\varepsilon = 0.5$    (b) $\varepsilon = 2$    (c) $\varepsilon = 5$    (d) $\varepsilon = 7$

Figure 2: $\ell_1$-error comparison between four algorithms $k = 1000$ and $p \sim U[k]$

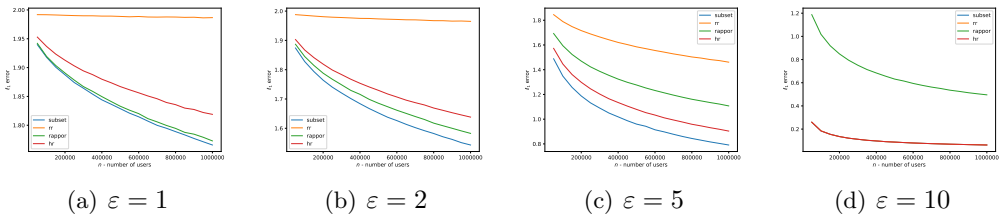

(a) $\varepsilon = 1$    (b) $\varepsilon = 2$    (c) $\varepsilon = 5$    (d) $\varepsilon = 10$

Figure 3: $\ell_1$-error comparison between four algorithms $k = 10000$ and $p \sim Geo(0.8)$

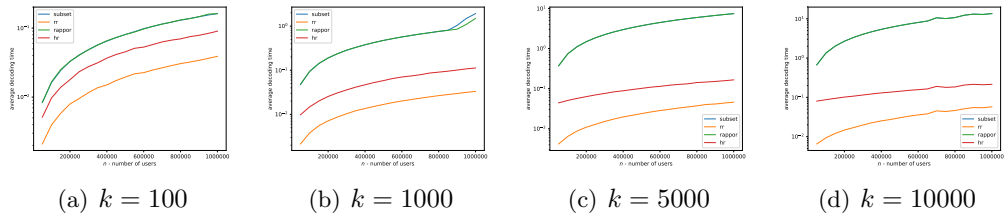

(a) $k = 100$    (b) $k = 1000$    (c) $k = 5000$    (d) $k = 10000$

Figure 4: Decoding time comparison between four algorithms for $\varepsilon = 1$ and $p \sim Geo(0.8)$ and different values of $k$

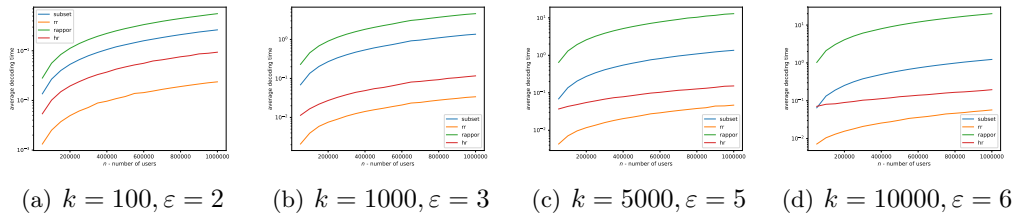

(a) $k = 100, \varepsilon = 2$    (b) $k = 1000, \varepsilon = 3$    (c) $k = 5000, \varepsilon = 5$    (d) $k = 10000, \varepsilon = 6$

Figure 5: Decoding time comparison between four algorithms in middle privacy regime and $p \sim Geo(0.8)$. Note that the decoding times are in logarithmic scale.

# A Proof of Lemma 6

Let $T_i = \{x \in [k] \mid r_x \text{ is in the } i\text{th } H_b \text{ block}\}$ be the set of symbols such that $r_x$ is in the $i$th $H_b$ block. From the description of $r_x$, we obtain

$$T_i := \left\{x \mid \lfloor \frac{x}{b-1} \rfloor = i \right\}, \quad p(T_i) := \sum_{x \in T_i} p(x) \text{ and, } \sum_i p(T_i) = 1. \tag{17}$$

We will prove that

$$p(C_x) = \frac{1}{2B - 1 + e^\varepsilon} + \frac{e^\varepsilon - 1}{2(2B - 1 + e^\varepsilon)} p(x) + \frac{e^\varepsilon - 1}{2(2B - 1 + e^\varepsilon)} p(T_i), \text{ and} \tag{18}$$

$$p(S_i) = \frac{e^\varepsilon - 1}{2B - 1 + e^\varepsilon} p(T_i) + \frac{2}{2B - 1 + e^\varepsilon} \tag{19}$$

Then note that (18)$-\frac{1}{2}$ (19) gives Lemma 6.

*Proof of* (18). Recall that for any $x$,

$$p(C_x) = \sum_{x'} p(x') Q(Z \in C_x | X = x'). \tag{20}$$

For any $x, x' \in [k]$, by (3) and $s = b/2$,

$$Q(Z \in C_x | X = x') = \frac{2e^\varepsilon}{be^\varepsilon + 2K - b} \times |C_x \cap C_{x'}| + \frac{2}{be^\varepsilon + 2K - b} \times |C_x \setminus C_{x'}|. \tag{21}$$

There are three cases:
- $x' = x$. In this case, $|C_x \cap C_{x'}| = s = b/2$, and $|C_x \setminus C_{x'}| = \emptyset$.
- $x' \in T_{\lfloor \frac{x}{b-1} \rfloor} \setminus \{x\}$: When this happens, then by the Property (ii) of Hadamard matrices, $|C_x \cap C_{x'}| = s/2 = b/4$, and $|C_x \setminus C_{x'}| = s/2 = b/4$.
- $x' \notin T_{\lfloor \frac{x}{b-1} \rfloor}$: The symbols $x' \notin T_{\lfloor \frac{x}{b-1} \rfloor}$ satisfy $|C_x \cap C_{x'}| = \emptyset$, and $|C_x \setminus C_{x'}| = b/2$.

Plugging these in (21), and using (20) with $K = Bb$, we obtain

$$p(C_x) = p(x) \frac{e^\varepsilon}{2B - 1 + e^\varepsilon} + p\big(T_{\lfloor \frac{x}{b-1} \rfloor} \setminus \{x\}\big) \frac{e^\varepsilon + 1}{2(2B - 1 + e^\varepsilon)} + p\big([k] \setminus T_{\lfloor \frac{x}{b-1} \rfloor}\big) \frac{1}{2B - 1 + e^\varepsilon}$$

$$= \frac{e^\varepsilon \cdot p(x)}{2B - 1 + e^\varepsilon} + \frac{e^\varepsilon + 1}{2(2B - 1 + e^\varepsilon)} \big(p(T_{\lfloor \frac{x}{b-1} \rfloor}) - p(x)\big) + \frac{1}{2B - 1 + e^\varepsilon} \big(1 - p(T_{\lfloor \frac{x}{b-1} \rfloor})\big)$$

$$= \frac{1}{2B - 1 + e^\varepsilon} + \frac{e^\varepsilon - 1}{2(2B - 1 + e^\varepsilon)} p(x) + \frac{e^\varepsilon - 1}{2(2B - 1 + e^\varepsilon)} p(T_{\lfloor \frac{x}{b-1} \rfloor}).$$

*Proof of* (19). Recall that $p(S_i)$ is the probability that the output is in $S_i$, when input distribution is $p$. Note that for $x \in T_i$, $C_x \subset S_i$, hence $|S_i \cap C_x| = |C_x| = b/2$ and $C_x \setminus S_i = \emptyset$. For $x \notin T_i$, $C_x \cap S_i = \emptyset$ and $|C_x \setminus S_i| = |C_x| = b/2$. Again using (21) and replacing $C_x$ with $S_i$, we obtain

$$p(S_i) = \frac{e^\varepsilon + 1}{2B - 1 + e^\varepsilon} p(T_i) + \frac{2}{2B - 1 + e^\varepsilon} (1 - p(T_i)).$$

Rearranging the terms gives (19).

# B  Proof of sample complexity bounds (Theorem 7)

We will prove Theorem 7 from Lemma 6. The proof follows the general approach used in Section 4 for the high privacy regime. Subtracting (13) of Lemma 6 from (16), we obtain

$$\hat{p}(x) - p(x) = \frac{2(2B - 1 + e^\varepsilon)}{e^\varepsilon - 1} \cdot \left( \left( \widehat{p(C_x)} - p(C_x) \right) - \frac{1}{2} \left( \widehat{p(S_i)} - p(S_i) \right) \right), \tag{22}$$

where recall that $S_i$ is the output columns corresponding to the $r_x$th block. Squaring both sides, and observing that for any reals $(a-b)^2 \le 2(a^2 + b^2)$, we obtain

$$(\hat{p}(x) - p(x))^2 \le \frac{8(2B - 1 + e^\varepsilon)^2}{(e^\varepsilon - 1)^2} \cdot \left( \left( \widehat{p(C_x)} - p(C_x) \right)^2 + \frac{1}{4} \left( \widehat{p(S_i)} - p(S_i) \right)^2 \right). \tag{23}$$

Now recall that $\widehat{p(C_x)}$ is average of independent Bernoulli's with mean $p(C_x)$, and $\widehat{p(S_i)}$ is the average of independent Bernoulli's with mean $p(S_i)$. Therefore,

$$\mathbb{E}\left[ \left( \widehat{p(C_x)} - p(C_x) \right)^2 \right] = \frac{1}{n} p(C_x)(1 - p(C_x)) < \frac{1}{n} p(C_x), \tag{24}$$

and

$$\mathbb{E}\left[ \left( \widehat{p(S_i)} - p(S_i) \right)^2 \right] = \frac{1}{n} p(S_i)(1 - p(S_i)) < \frac{1}{n} p(S_i). \tag{25}$$

Summing over $x$ in (18) and using (17), we obtain

$$\sum_x \mathbb{E}\left[ \left( \widehat{p(C_x)} - p(C_x) \right)^2 \right] < \frac{1}{n} \sum_x \left( \frac{1}{2B - 1 + e^\varepsilon} + \frac{e^\varepsilon - 1}{2(2B - 1 + e^\varepsilon)} p(x) + \frac{e^\varepsilon - 1}{2(2B - 1 + e^\varepsilon)} p(T_i) \right)$$

$$\le \frac{1}{n} \left( \frac{k}{2B - 1 + e^\varepsilon} + \frac{e^\varepsilon - 1}{2(2B - 1 + e^\varepsilon)} + \frac{b(e^\varepsilon - 1)}{2(2B - 1 + e^\varepsilon)} \right). \tag{26}$$

where the last inequality follows since each $T_i$ is of size at most $b$, and $\sum_i p(T_i) = 1$, implying that $\sum_x p(T_i) \le b$. Similarly, summing over $x$ in (19),

$$\sum_x p(S_i) \le \frac{b(e^\varepsilon - 1)}{2B - 1 + e^\varepsilon} + \frac{2k}{2B - 1 + e^\varepsilon}. \tag{27}$$

Summing over $x$ in (23) and taking the expectations, and plugging the bounds above with the observation that $B < e^\varepsilon$ by design, we obtain the bound on $\mathbb{E}\left[\ell_2(\hat{p}, p)^2\right]$, proving the theorem.

$$\mathbb{E}\left[\ell_2^2(\hat{p}, p)\right] \le \frac{1}{n} \frac{8(2B - 1 + e^\varepsilon)^2}{(e^\varepsilon - 1)^2} \left( \left( \frac{2k + (b+1)(e^\varepsilon - 1)}{2(2B - 1 + e^\varepsilon)} \right) + \frac{1}{4} \left( \frac{4k + 2b(e^\varepsilon - 1)}{2(2B - 1 + e^\varepsilon)} \right) \right)$$

$$= \frac{1}{n} \frac{4(2B - 1 + e^\varepsilon)}{(e^\varepsilon - 1)^2} \left( 3k + (\frac{3}{2}b + 1)(e^\varepsilon - 1) \right). \tag{28}$$

$$\le \frac{36 e^\varepsilon (k + b(e^\varepsilon - 1))}{n(e^\varepsilon - 1)^2}.$$

# C Description and performance of RAPPOR and SS

## C.1 $k$-RAPPOR.

Recall from Section 2.2 the privatization mechanism of RAPPOR. For input $x \in [k]$, $\mathbf{y} \in \{0,1\}^k$ is such that $\mathbf{y}_j = 1$ for $j = x$, and $\mathbf{y}_j = 0$ for $j \neq x$. The privatized output of RAPPOR is a $k$ bit vector $\mathbf{z}$ such that

$$Q(\mathbf{z}_j = \mathbf{y}_j) = \frac{e^{\varepsilon/2}}{e^{\varepsilon/2} + 1}, \text{ and } Q(\mathbf{z}_j = 1 - \mathbf{y}_j) = \frac{1}{e^{\varepsilon/2} + 1}.$$

[31] analyze the sample complexity of RAPPOR (See Table 1). We will consider the communication requirements now in Theorem 9.

**Communication.** The output of $k$-RAPPOR mechanism is described above with $k$ bits. We now consider the communication requirements for any algorithm that faithfully sends the output of RAPPOR privatization to the server. By Shannon's coding theorem, any algorithm to do this requires at least $H(Z|p)$ bits even if it knows the distribution $p$.

**Theorem 9.** *The entropy $Z$ of the output of RAPPOR for any input distribution satisfies*

$$H(Z) \geq \begin{cases} \Omega(k) & \text{when } \varepsilon < 1, \\ \Omega\left(\frac{k}{e^{\varepsilon/2}}\right) & \text{when } 1 < \varepsilon < 2\log k, \end{cases}$$

*and for the uniform input distribution $H(Z) \geq \log k$ when $\varepsilon > 2\log k$.*

*Proof.* For any input $x$, the outputs $\mathbf{z}_j$ for $j \neq x$ are all i.i.d. $B(\frac{1}{1+e^{\varepsilon/2}})$ random variables, where $B(r)$ is a Bernoulli random variable with bias $r$. Therefore the entropy of the output is at least $(k-1) \cdot h(1/(1+e^{\varepsilon/2}))$, where $h(r) := -r\log r - (1-r)\log(1-r)$ is the entropy of a $B(r)$ random variable. Note that $h(r) > -r\log r$. Therefore, $(k-1)h(1/(1+e^{\varepsilon/2})) > (k-1)\frac{\log(1+e^{\varepsilon/2})}{1+e^{\varepsilon/2}}$. For $\varepsilon < 1$ this bound reduces to $\Omega(k)$, and for any $\varepsilon < 2\log k$ ignoring the logarithmic term gives the theorem. For the uniform input distribution, when $\varepsilon > 2\log k$, we note that the output is nearly uniform on the basis vectors, giving the bound. □

## C.2 Subset Selection Approaches.

The papers [43, 47] propose sample optimal privacy mechanisms for all ranges of $\varepsilon$. The mechanism is as follows. The output is again $k$ bits, and suppose $d = \lceil k/(e^\varepsilon + 1) \rceil$. The output is $\mathcal{Z} = \mathcal{Z}_{k,d}$, where $\mathcal{Z}_{k,d}$ is the set of all binary strings with Hamming weight $d$. For an $i \in [k]$, let $\mathcal{Z}^i_{k,d}$ be the elements in $\mathcal{Z}_{k,d}$ with 1 in the $i$th location. Then note that $|\mathcal{Z}_{k,d}| = \binom{k}{d}$, and $\left|\mathcal{Z}^i_{k,d}\right| = \binom{k-1}{d-1}$. Then, for $Z_1 \ldots Z_k \in \mathcal{Z}_{k,d}$,

$$Q(Z_1 \ldots Z_k | i) = \begin{cases} \frac{e^\varepsilon}{\binom{k-1}{d-1}e^\varepsilon + \binom{k-1}{d}}, & \text{for } Z_1 \ldots Z_k \in \mathcal{Z}^i_{k,d}, \\ \frac{1}{\binom{k-1}{d-1}e^\varepsilon + \binom{k-1}{d}}, & \text{for } Z_1 \ldots Z_k \in \mathcal{Z}_{k,d} \setminus \mathcal{Z}^i_{k,d}. \end{cases}$$

**Communication.** We can characterize the communication complexity by computing the entropy of the output distributions of the mechanism. The computations are similar to the last section, we simply state the entropy bounds that imply the communication bounds.

Suppose the underlying distribution is uniform. In this case, the output distribution is uniform among all possible strings of weight equal to $d$. Therefore the entropy of the output string is identical to $\log \binom{k}{d}$, which is the optimal communication complexity per user. Note that for small $\varepsilon$ this communication is strictly undesirable!

**Theorem 10.** *The entropy $Z$ of the output of SS for any input distribution satisfies*

$$H(Z) \geq \begin{cases} \Omega(k) & \text{when } \varepsilon < 1, \\ \Omega\left(\frac{k}{e^\varepsilon}\right) & \text{when } 1 < \varepsilon < \log k, \end{cases}$$

*and for the uniform input distribution $H(Z) \geq \log k$ when $\varepsilon > \log k$.*

While the final mechanism is sample order optimal for all values of $\varepsilon$, the communication cost for each user depends critically on the value of $\varepsilon$. Table 2 characterizes the communication cost for all three well known mechanisms and our proposed method.